# A Formal Framework for the Definition of 'State': Hierarchical Representation and Meta-Universe Interpretation

Kei Itoh*

## Abstract

This study aims to reinforce the theoretical foundation for diverse systems—including the axiomatic definition of intelligence—by introducing a mathematically rigorous and unified formal structure for the concept of "state," which has long been used without consensus or formal clarity. First, a "hierarchical state grid" composed of two axes—state depth and mapping hierarchy—is proposed to provide a unified notational system applicable across mathematical, physical, and linguistic domains. Next, the "Intermediate Meta-Universe (IMU)" is introduced to enable explicit descriptions of definers (ourselves) and the languages we use, thereby allowing conscious meta-level operations while avoiding self-reference and logical inconsistency. Building on this meta-theoretical foundation, this study expands inter-universal theory beyond mathematics to include linguistic translation and agent integration, introducing the conceptual division between macrocosm-inter-universal and microcosm-inter-universal operations for broader expressivity.

Key contributions include:

1. Formalization of the previously ambiguous, domain-incompatible concept of "state," integrating definition, state, and structure under a unified concept.
2. Encoding of all states—across physics, logic, and language—onto the (state depth, mapping hierarchy) coordinate system of the hierarchical state grid.
3. Axiomatization of "definability" at the lowest layer of the grid to systematically accommodate unobserved or future phenomena.
4. Structural visualization of predicates, propositions, and programs, making their dependency structure explicitly observable.
5. Booleanization of intelligence judgment functions, enabling their direct application to AI implementation.
6. Extension of the grid into a time dimension, separating virtual time from real time with formal precision.
7. Introduction of the IMU to isolate self-referential contradictions, allowing definers, definition languages, and meta-definitions to be treated as objects.

8. Implementation of translation, integration, and real-time mappings within the IMU to coherently

manage multilingualities, multi-agent definitions, and temporal change.

9. Bifurcation of inter-universal operations into macrocosmic and microcosmic classes, thereby enabling application to non-mathematical domains.
10. Clarification of the distinction between proof (static validation) and verification (dynamic validation) via temporal coordinates.
11. Establishment of a codomain requirement: all definitions must eventually be transported into verifiable universes, thus formalizing the obligation of validity.

Through these contributions, this paper presents a meta-formal logical framework—grounded in the principle of *definition* = *state*—that spans time, language, agents, and operations, providing a mathematically robust foundation applicable to the definition of intelligence, formal logic, and scientific theory at large.

* Ehime University Graduate School of Science and Technology. The author conducted this research independently while affiliated with Ehime University and utilized the university's library and electronic resources. The research was conducted without specific financial support from external funding sources. The author declares no conflict of interest. While this English version constitutes the formal and citable edition of the work, the original manuscript was composed in Japanese, and some structural or rhetorical nuances may be more fully preserved in the original text.

## Introduction

Itoh (2025) [1] attempted to provide an axiomatic definition of "intelligence" by constructing a set-theoretic or category-theoretic universe. This paper implemented a protocol that makes it possible to introduce mathematical logic into initially naive concepts. It carried out various axiomatic definitions of intelligence based on set-theoretic logic. In this context, various concepts grounded in set-theoretic logic—such as "element" and "(time) mapping"—are introduced into the definition of intelligence. However, among these, the concept of "state" is not explicitly defined or operationalized in the paper. Furthermore, even within basic set-theoretic logic itself, the concept of "state" continues to be used as an ambiguous term, with no explicit and commonly agreed-upon definition. "State" is defined differently and non-isomorphically across fields: as real-valued vectors in control theory [e.g., 2], as positive linear functionals in C*-algebras [e.g., 3], as morphisms in monoidal categories [e.g., 4], and as carrier set elements in coalgebras [e.g., 5]. This ambiguity surrounding the notion of state constitutes a critical defect in axiomatic definitions of intelligence, which demand explicit and mathematically rigorous operations on concepts. Therefore, this paper aims first to bridge the definitional disconnect of the concept of "state" across these fields through reinterpretation, and to reinforce the theoretical foundation for the formal definition of intelligence. To this end, a "hierarchical state grid" is introduced, where the horizontal axis represents state depth. The vertical axis represents mapping hierarchy, allowing any and all entities that may be referred to as "state"—be they numerical, physical, or belonging to any other domain—to be uniformly represented within a single formalism. In this construction, the concept of "definition" itself can be regarded as a specific type of state, namely a "definitional mapping," which shows that the "determination of state" and the "determination of definition" for a given object are, in essence, the same operation. Through this development, two of the most logically fundamental concepts— "state" and "definition" —are unified. Furthermore, by defining state in hierarchical terms, it becomes clear that the concept of "time" splits into two distinct conceptual entities: "virtual time," which is introduced as a temporal state within a given definition, and "real time," which continuously acts upon ourselves. In other words, virtual time is a non-real, numerically structured time that is representable through a particular formal notation and thus directly definable. In contrast, real time is a meta-level and highly abstract concept that, for reasons discussed later, cannot be entirely and directly defined. This distinction is made clear through the introduction of the hierarchical state model.

In order to make the above construction more comprehensive and smoothly operable, this paper explicitly and consciously introduces meta-conceptual entities such as "definer," "definition language," and "definition of definition" into this theory of definitions. First, a hierarchical distinction is made between the truly meta-level universe in which we ourselves exist and the universe that serves as the object of definition or contains definitions. The former is referred to as the upper level (true meta-universe), and the latter as the lower level (definition universe). In this study, the term "universe" is used in the sense of a mathematical universe, but not in the sense of a specific construction like the Grothendieck universe [e.g., 6]. Instead, it is used more broadly to refer to the set of all objects that a given axiomatic system can describe. Here, the complete definition of ourselves as an object is clearly impossible, either due to formal limitations such as Gödel's incompleteness theorem [7] or Russell's paradox [8], or due to physical considerations such as the Pauli exclusion principle [9], which implies that we ourselves cannot fully describe ourselves directly. Therefore, this study introduces a "intermediate meta-universe" as a representational buffer layer between the true meta-universe and the definition universe. The intermediate meta-universe is constructed from mirror-like objects formed through self-mapping projections of entities existing in the true meta-universe, and meta-concepts within the definition universe are defined through morphisms originating from objects in the intermediate meta-universe. In this way, the construction of a meta-theory makes it possible to handle meta-concepts in definitions explicitly, while avoiding the impossibility of complete definition as imposed by the incompleteness theorems. Furthermore, this approach also provides a methodology for addressing the previously mentioned issue of how to handle real-time.

This meta-definition theory strongly motivates the introduction of methods for operating among different definition universes, each characterized by the distinct definition languages used in Itoh (2025)—such as the "naive definition," "set-theoretic definition," and "category-theoretic definition" of intelligence. In other words, the assertion that "any language may be used as long as practical validity is preserved" essentially demands the explicit introduction of inter-universal operational methods. The term "inter-universal operation" here refers to mathematical-logical transformations across universes (≈ axiomatic systems), such as size control and model conversion of categories as in the "Change of Universe" from homotopy theory [e.g., 10], or logical constructions of structure deformation across universes as in Inter-Universal Teichmüller Theory (IUT) [11]. Motivated by Itoh (2025), this study aims to introduce a mathematical-logical framework that encompasses inter-universal operations not only among mathematical languages but also among all conceivable universes, including those not described in mathematical terms. That is, the goal is to expand the initially mathematically confined notion of inter-universality into a universally applicable principle across the sciences in general. Furthermore, this development simultaneously demonstrates that the "codomain" of inter-linguistic inter-universal transformations, such as the definition of intelligence in Itoh (2025), must be a universe in which the truth or practical validity of that definition can be verified within real-time progression. From this, the following assertions naturally arise: that mathematical logic with ambiguous proof completion should be inter-universally transformed into a language suitable for machine verification in order to validate proof completion; and that axiomatizations of naive concepts should ultimately be transformed into languages that permit mechanical implementation (e.g., programming languages), allowing for verification of their practical utility.

Accordingly, the objective of this study is to construct the following three logical frameworks:

1. A hierarchical state grid — a coordinate system in which state and definition are treated under a unified format

2. A meta-theory focused on the intermediate meta-universe — a buffer layer for externalizing definition operations while avoiding the limitations of self-description

3. An inter-universal theory within and outside the definition universe — a procedure for transporting definitions across languages and axiomatic systems to a codomain where verification is possible

In section 1, in addition to the above, the concrete method of defining and operating the hierarchical state grid will be demonstrated through examples such as the function continuity judgment mapping and the axiomatic definition of intelligence given in Itoh (2025). Both natural language-based set-theoretic constructions and formal definitions using ZFC-based first-order predicate logic will be described. Moreover, since the concepts of state and definition are strictly formalized, the inadequacy in the definition of "structure" in Itoh (2025) becomes evident, and a clear reformulation will be provided. Throughout sections 1–3, the theory of state and its meta-theoretical interpretation with respect to the concept of time will be included. As a secondary consequence of this, it will be shown that the concept of "proof" in mathematics does not necessarily guarantee logical correctness in non-mathematical domains. In non-mathematical fields, there are effectively no operations whose correctness is assured across all time; rather, the validity of a theory can be guaranteed only through "verification" within a finite time window.

### State Definition Theory by Introduction of the "Hierarchical State Grid"

The "hierarchical state grid" constructed in this study is a notational system that allows for the inclusion and definition of various types of "states" by taking the horizontal axis as state depth and the vertical axis as mapping hierarchy. State depth refers to the level difference in which a higher-order state is defined based on a lower-order one. In any scientific field, the combination of simpler states generally determines a complex state, and this concept is introduced to organize and represent such relationships in a unified manner. On the other hand, mapping hierarchy refers to whether the domain of the conceptual mapping being defined is itself a mapping, and indicates its level or order (for example, as in higher category theory). A 0th-order state is a non-mapping, label-like state; a 1st-order mapping is a mapping whose domain is such a label-like state; a 2nd-order mapping is a mapping whose domain is a 1st-order mapping, and so forth. As the hierarchy rises, the order of the mapping state increases. Table 1 shows a conceptual configuration of the hierarchical state grid, and Table 2 presents a concrete example of grid construction using numerical system states. From Table 1, it is clear that the grid is coordinate-based using Cartesian notation. Each coordinate in this grid represents the configuration of a particular state.

| Mapping hierarchy | | | | | | | |
|---|---|---|---|---|---|---|---|
| ⋮ | ⋮ | ⋮ | ⋮ | ⋮ | ⋮ | | |
| 3rd-order mappings | (0,3) | (1,3) | (2,3) | (3,3) | (4,3) | … | |
| 2nd-order mappings | (0,2) | (1,2) | (2,2) | (3,2) | (4,2) | … | |
| 1st-order mappings | (0,1) | (1,1) | (2,1) | (3,1) | (4,1) | … | |
| 0-order states | (0,0) | (1,0) | (2,0) | (3,0) | (4,0) | … | |
| | Depth 0 | Depth 1 | Depth 2 | Depth 3 | Depth 4 | … | State Depth |

**Table 1.** Conceptual structure of the hierarchical state grid. Each coordinate of the grid represents (horizontal, vertical) = (state depth, mapping hierarchy).

| Mapping hierarchy | Depth 0 | Depth 1 | Depth 2 | Depth 3 | Depth 4 | Depth 5 |
|---|---|---|---|---|---|---|
| 3rd-order mappings | (0,3) | (1,3) Higher-order composite truth function | (2,3) Higher-order cardinality properties | (3,3) Higher-order order properties | (4,3) Higher-order field properties | (5,3) Higher-order continuum properties |
| | | | bijectivity test, infinite cardinality test | induction predicates | algebraic closure test, continuity of field operations | intermediate-value theorem, Banach-space test |
| 2nd-order mappings | (0,2) | (1,2) Composite truth function | (2,2) Cardinality predicates | (3,2) Order predicates | (4,2) Field-property predicates | (5,2) Continuity predicates |
| | | logical OR/AND | cardinality-equality / isomorphism / countability | max/min element | topologisability, metrisability of a field | continuity / differentiability test |
| 1st-order mappings | (0,1) Definability-test mapping | (1,1) Truth-value function | (2,1) Cardinality mapping | (3,1) Order-structure mapping | (4,1) Field/operation mapping | (5,1) Continuity mapping |
| | | logic gates | countable / uncountable test | order + Peano addition | four basic operations | complete functions |
| 0-order states | (0,0) Definability | (1,0) Empty / Non-empty | (2,0) Base set | (3,0) Ordered set | (4,0) Field / Ring set | (5,0) Complete field |
| | State of definability (meta-Boolean) | Boolean | unordered basis | $\mathbb{N}$, $\mathbb{Z}$, total/partial order | $\mathbb{Q}$, algebraic number system | $\mathbb{R}$, complete continuum |
| State Depth | Depth 0 | Depth 1 | Depth 2 | Depth 3 | Depth 4 | Depth 5 |

**Table 2.** Representation of numerical system states using the hierarchical state grid. The deeper the state depth, the more complex the conceptual state; the higher the mapping hierarchy, the more predicate-like the state. Mapping hierarchy is likely to be isomorphically determined for any mapping concept, but state depth may vary arbitrarily even within the same conceptual state system.

As shown in Table 2, the introduction of the hierarchical state grid allows for simultaneous representation of both low- and high-order states and their mapping orders. Since neither state depth nor mapping hierarchy restricts their content to mathematical, physical, or other specific domains, this notation can incorporate and define state concepts from any field, making cross-domain comparison possible. Moreover, because this is a purely notational system, the specific contents of state depth and mapping hierarchy will vary depending on the states one wishes to define. What is particularly important here is the treatment of not only label-like non-mapping states as states, but also mapping concepts themselves as states, thus unifying label concepts and mapping concepts under the notion of "state." In doing so, not only can predicates be regarded as states, but even mathematical systems such as function spaces that inherently contain mapping behavior can also be included as state concepts. Moreover, because this construction treats not only label-like states but also any higher-order predicate as a single state, in this notation system, the "determination of a definition" and the "determination of a state" are identified as the same operation, thereby unifying the concepts of definition and state.

In Table 2, the state corresponding to depth zero is designated as "definability" or its negation, representing a fundamental state that is axiomatized within the act of definition itself—namely, whether the object to be defined is in fact definable. Since something becomes definable the moment it is defined, unless a meta-perspective is introduced, every state or definition would trivially have the property of "definable." Thus, it is considered appropriate to introduce "definability" as a fundamental state. That is, a Boolean value at depth 0 represents an ontological meta-evaluation of whether a given object or state is "definable" within our descriptive system; even the "empty set" counts as a definable state once it is defined as such. In contrast, a Boolean value at depth 1 applies to already defined objects, evaluating the truth or falsity of their attributes or predicates—e.g., whether something is empty or not, whether a property holds or not. From this description alone, depth zero may appear to be an unnecessary axiom, but as will be discussed later, this depth distinction becomes meaningful when real time is taken into consideration. In the progression of real time, we inevitably encounter states or predicates that are genuinely "undefinable." There certainly exist past states that are no longer accessible; the set of states currently knowable is limited; and future states cannot be known in the present. Thus, when we take into account real-world "real time," unobserved, unknowable, or future events must be judged as "undefinable" at depth 0, and the theoretical domain of such states can become infinitely large. In this way, when the introduction of real time into the theory of definition is considered, definability becomes the shallowest state. It should be axiomatically introduced into virtually any state concept.

**Concrete State Definition Using the Hierarchical State Grid**

**Definition of the Continuity Judgment Mapping**

As an example of a concrete state definition using the hierarchical state grid, we first define a continuity judgment mapping for a function. The following mathematical expressions are written using first-order predicates under the ZFC system.

First, define the function mapping $f: \mathbb{R} \to \mathbb{R}$ as:

$$\mathrm{Func}(f) \equiv \forall x \forall y \forall z\big((x,y) \in f \land (x,z) \in f \to y = z\big), (\, x, y, z \in \mathbb{R}).$$

Next, define the continuity predicate $\varphi_{\mathrm{Cont}}(f)$ as:

$$\varphi_{\mathrm{Cont}}(f) \equiv \mathrm{Func}(f) \land$$
$$\forall a \in \mathbb{R} \forall \varepsilon \in \mathbb{R}^{+}$$
$$\exists \delta \in \mathbb{R}^{+} \forall x \in \mathbb{R}(|x - a| < \delta \to |f(x) - f(a)| < \varepsilon)^{(\mathrm{C})}.$$

Here,

$$|x - a| < \delta \equiv (x - a < \delta) \lor (a - x < \delta)$$

can be expanded using the order relation $<$ and field operations such as $+, -, |\cdot|$. Then, we transform $\varphi_{\mathrm{Cont}}(f)$ into a truth-evaluating function $\mathrm{Cont}(f)$, a Boolean judgment mapping:

$$\mathrm{Cont}(f) \equiv \begin{cases} 1 & \text{if } \varphi_{\mathrm{Cont}}(f) \text{ holds true} \\ 0 & \text{otherwise} \end{cases}$$

Here, 0 = False, and 1 = True.

As shown in Table 3, all of the mathematical expressions used for this definition can be positioned on the grid in Table 2. As demonstrated here, when the entire content of a definition is written out as a state hierarchy, all the involved states are made explicit, and the internal structure becomes manifest. By expressing every component of a definition explicitly on the hierarchical state grid, the essential structure, dependency relations, and operational depth of that definition are rendered visible.

Here, in propositional logic, propositions and their truth values are fundamentally expressed as combinations of truth value functions, and in computer science, any description of predicates is likewise composed through combinations of logic gates. From this, it becomes evident that the hierarchical state grid discussed thus far—especially about mathematical or information-scientific

predicates—allows any state depth or mapping height to be unfolded through the combination of the (1,1) truth value function and its domain at (1,0), i.e., the empty/non-empty state.

| Grid Coord. | Label | Part Appearing in Formula (C) |
|---|---|---|
| (5,0) | Real-number field $\mathbb{R}$ | Domain of variables $a, x, \varepsilon, \delta$ |
| (4,1) | Field operations $+, -, \vert\cdot\vert$ | Arithmetic in $x - a,\ f(x) - f(a)$ |
| (3,2) | Order predicate "<" | Inequalities |
| (5,1) | Function mapping $f\colon \mathbb{R} \to \mathbb{R}$ | $f$ as a predicate indicating a function mapping |
| (5,2) | $\varphi_{\mathtt{Cont}}(f)$ | Definition of continuity |
| (5,3) | $\mathtt{Cont}(f)$ | Boolean test of continuity |
| (1,0) | Truth values {True, False} | Result of the continuity test |

**Table 3.** Placement of each component of the continuity judgment mapping on the hierarchical state grid (corresponding to Table 2).

**Reinterpretation of the Axiomatic Definition of Intelligence by Itoh, 2025**

Next, as another example of a concrete definition, we reinterpret the axiomatic definition of intelligence given in Itoh (2025) using the hierarchical state grid. To do so, we first redefine the terms "element," "existence," and "structure" as used in Itoh (2025), introducing the concept of state as made explicit in this paper.

Let the element $\upsilon$ of the universal set $U$ be considered as a state s:

$$U \coloneqq \{\upsilon|\text{Universe}\} \coloneqq \{s|\text{Universe}\}$$

In other words, the universe is regarded as the set of all states existing within it. That an existence $E$ within the universe is a subset of $U$ (i.e., $E \subseteq U$) remains unchanged from Itoh (2025), but due to the introduction of the state concept in this study, this is redefined as a set of states instead of a set of elements. Then, a structure $C$ in an existence $E$ is not represented by a Cartesian product set, but is instead regarded as a specific state within E. This means that there is no longer a need to define the concept of "structure" separately. Within the hierarchical state grid, any predicate, relational expression, or mapping-based construction can be represented as a state at a specific coordinate, and this is regarded as a "predicate state." Therefore, just as "definition" and "state" were identified, "structure" and "state" are also identified. In practical terms, a structure is presented as a state at a particular coordinate, and its definition will depend on the specific nature of that definition. Thus, under the reinterpretation presented in this study, a structure is nothing more than a label attached to a "predicate state" appearing at a higher-order coordinate. This interpretation implies, for example, that Cartesian products are merely convenient tools for generating such structural states and are not essential components (Similarly, power sets or the concepts of absolute and relative state may often be used in defining structural states.). Therefore, this paper demonstrates that the terms "definition," "state," and "structure" are entirely interchangeable names for the same concept, enabling the concept of "structure" in any field to be integrated under a unified representation.

Next, the axiomatic definition of intelligence in Itoh (2025) is mapped onto the hierarchical state grid and redefined as a truth-evaluating predicate. The idea that time progression is conceptualized as a mapping, and the naive definition that "An entity possessing structures for receiving information or matter from the external environment (input structure), processing it internally (processing structure), and outputting it externally (output structure)," as well as the predicates based on cardinality and temporal order operations, are adopted as is. The input structure $C_i$, output structure $C_o$, and processing structure $C_p$ are respectively expressed in ZFC-based first-order predicate logic as:

$$C_i \coloneqq |I_{i+1}| > |I_i| \wedge |O_{i+1}| > |O_i|$$
$$C_o \coloneqq |I_{i+1}| < |I_i| \wedge |O_{i+1}| > |O_i|$$
$$C_p \coloneqq \exists T, V \subseteq I(|T_{i+1}| < |T_i| \vee |V_{i+1}| > |V_i|)$$

These are then transformed into Boolean functions $\mathrm{In}(I,O,i), \mathrm{Out}(I,O,i), \mathrm{Proc}(I,i)$ as follows:

$$\mathrm{In}(I,O,i) \equiv \begin{cases} 1 & \text{if } C_i \\ 0 & \text{otherwise} \end{cases}$$

$$\mathrm{Out}(I,O,i) \equiv \begin{cases} 1 & \text{if } C_o \\ 0 & \text{otherwise} \end{cases}$$

$$\mathrm{Proc}(I,i) \equiv \begin{cases} 1 & \text{if } C_p \\ 0 & \text{otherwise} \end{cases}$$

Here, 0 = False, and 1 = True.

Finally, these are integrated into an intelligence judgment mapping $\mathrm{Int}(I,O,i)$ as:

$$\mathrm{Int}(I,O,i) \equiv \begin{cases} 1 & \text{if } \mathrm{In}(I,O,i) = \mathrm{True} \wedge \mathrm{Out}(I,O,i) = \mathrm{True} \wedge \mathrm{Proc}(I,i) = \mathrm{True} \\ 0 & \text{otherwise} \end{cases}$$

Thus, the axiomatic definition of intelligence in Itoh (2025) is redefined as the Boolean function $\mathrm{Int}(I,O,i)$. The operations used here are limited to cardinality and order operations, and each component is placed on the hierarchical state grid as shown in Table 4. In this way, it is demonstrated that not only purely mathematical concepts, but also non-mathematical concepts such as intelligence, can all be reduced to higher-order predicate states and be definable within the hierarchical state grid.

| Grid Coord. | Label | Corresponding Component |
|---|---|---|
| (2,0) | Base set | $I, O$ |
| (2,1) | Cardinality mapping | $\|I\|, \|O\|$ |
| (3,0) | Time-series ordered set | $i, i+1$ |
| (3,1) | Order mapping (id, succ) | Time mapping $T_i: U_i \to U_{i+1}$ |
| (3,2) | Order predicates | Inequalities $\|I_{i+1}\| > \|I_i\|$ |
| (3,3) | $C_i, C_o, C_p$ | Intelligence-structure predicate |
| (3,4) | $\mathrm{In}, \mathrm{Out}, \mathrm{Proc}$ | Intelligence-structure test mapping |
| (3,5) | $\mathrm{Int}$ | Intelligence judgment mapping |
| (1.0) | Truth values {True, False} | Result of the intelligence judgment |

**Table 4.** Placement of each component of the intelligence judgment mapping on the hierarchical state grid (corresponding to Table 2).

**Extension of the Time Concept in the Hierarchical State Grid**

In Itoh (2025), time is defined as a time mapping, and as described above, it can indeed be expressed as a specific mapping state. However, treating the concept of time as a mapping is essentially a substitute method; in reality, time is a meta-level concept that continuously acts upon us and should, in a strict or practical sense, be considered as existing at a higher level than the very entities we define. This idea can be expressed as an extension of the hierarchical state grid by assigning the depth axis of the coordinate space to represent time (see Figure 1). That is, the original two-dimensional coordinate system (horizontal, vertical) = (state depth, mapping hierarchy) is extended to a three-dimensional coordinate system (horizontal, vertical, depth) = (state depth, mapping hierarchy, time). In this way, the concept of time in Itoh (2025) is no longer treated as a mapping, but is instead reduced to a coordinate within the hierarchical state grid. Rather than defining intelligence using a time mapping, the intelligence judgment mapping is constructed by comparing the contents of intelligent entities at different time coordinates. In such a construction, some components of Table 4 become unnecessary, and the progression of time is explicitly represented on the grid, allowing for a more organized state definition. Treating time as an independent dimension is equivalent to handling the time concept in a manner isomorphic to real time. It leads to a conceptualization of the meta-level operations of time discussed later, such as the distinction between the acts of “proof” and “verification.”

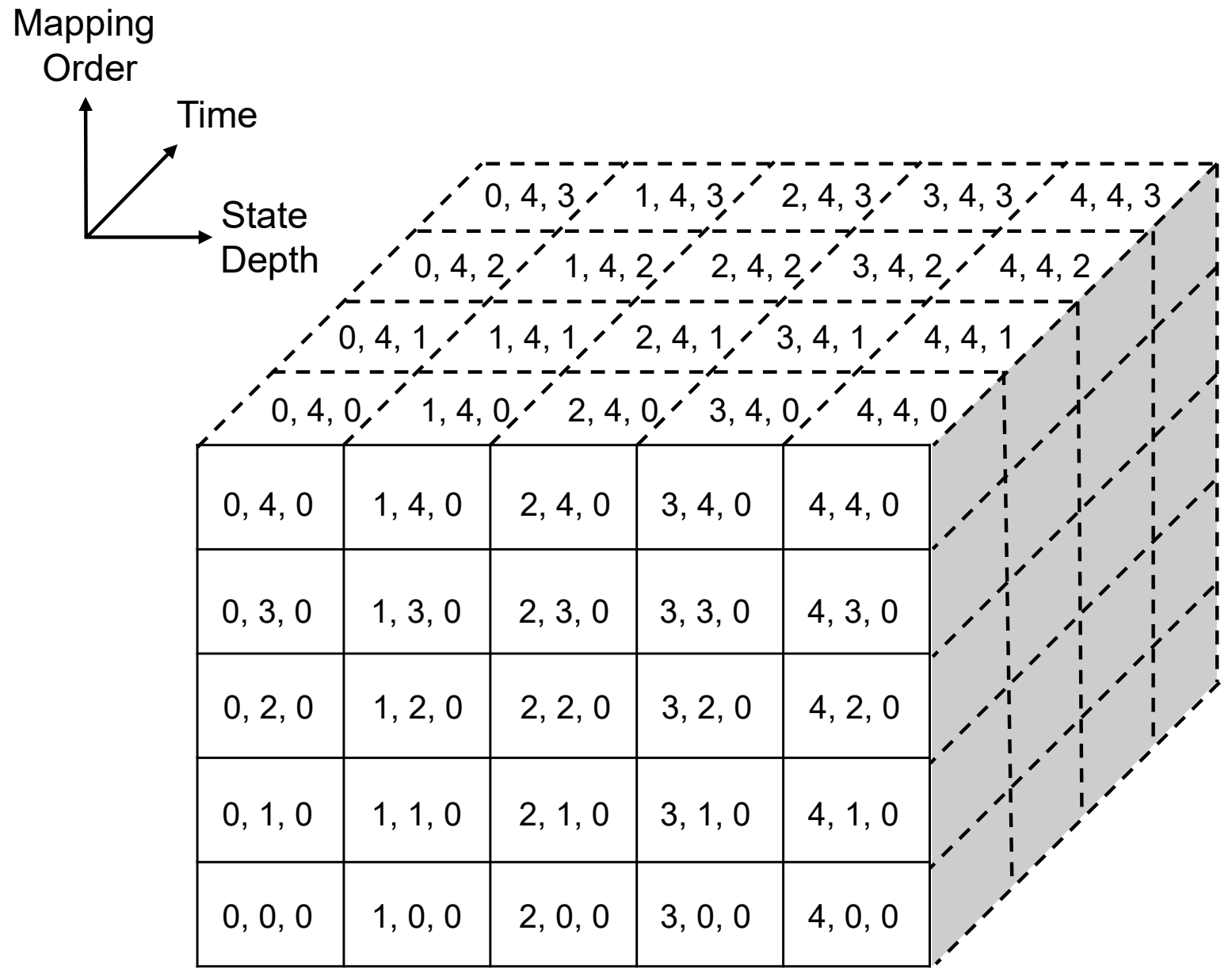

**Figure 1.** Conceptual structure of the time-extended hierarchical state grid. Each coordinate of the grid is represented as (horizontal, vertical, depth) = (state depth, mapping hierarchy, time).

**Meta-Theory on Definitions Centered on the Intermediate Meta-Universe**

This section details the essential significance and function of the concept of the "Intermediate Meta-Universe" (IMU), as well as its theoretical and practical roles. Traditionally, in any theoretical system concerning definitions, truly meta-level entities—such as "ourselves" as observers, the language and concepts we use, and the real time acting upon us—have been implicitly treated as operations. Likewise, a two-layer structure composed of non-meta descriptions (definition universe) such as symbol systems, axiomatic systems, and programming languages has also been tacitly assumed. This is presumably because any direct influence or recursive mapping from the true meta-universe to the definition universe would immediately trigger Gödelian incompleteness or self-referential paradoxes, thereby introducing fundamental risks to the soundness and consistency of the theoretical system itself.

To avoid this limitation of self-reference, it is considered essential to introduce a buffer layer, which is the concept of the "Intermediate Meta-Universe" (IMU) proposed in this study. The IMU stands between the true meta-universe and the definition universe as a set of mirror-like objects created by self-mapping projections of entities within the true meta-universe. Meta-level concepts within the definition universe (such as definer, definition language, and definition of definition) are then regarded as morphisms from the IMU. Furthermore, by extending this concept, it is shown that the IMU functions as a space that can safely handle all higher-order recursive operations, inter-universal mappings, and various forms of information transformation and integration. The essence of this layer lies in its ability to fundamentally block self-referential contradictions and formal breakdowns, while simultaneously providing an extremely flexible and powerful "external operation API" to the lower-level universe. At the implementation level, this can correspond to various meta-languages, interactive theorem-proving languages [e.g., 12,13], and multi-layered universe structures in type theory [e.g., 14,15]. That is, this concept has already been implicitly incorporated as the basis for programming language definitions, especially in computer science, and it is expected to serve as an expanded mechanism that generalizes such meta-concepts to the scientific domain at large.

The meta-concepts targeted by the IMU can, for example, be broadly divided into the following three categories.

**1. Definition Language (Figure 2)**

In general, it is universally observed that, within a given definition, different definition universes with varying syntax and semantics coexist simultaneously, such as set-theoretic logic, category-theoretic logic, and various programming or natural languages. Even when trying to describe the same phenomenon or structure across these universes, notations and definitions form differ significantly, and no consensus exists regarding their integration. As a result, implementations such as the intelligence definition in Itoh (2025) are currently feasible only as non-formal, artisanal methods. Here, the IMU provides the concept of a "translation mapping." In other words, the IMU treats the "words" and "grammar" of each language as its objects and represents the correspondences between languages as mappings. By using these, translation relations between definition universes can be formalized. For example, a "state" defined in set-theoretic logic could be transferred into category theory or another language as a mathematical logic object. If translation fails or incompatibility arises, this is explicitly marked as an "undefinable" definition, allowing ambiguity and incompleteness to be directly tracked within the theory. This can be regarded as a typical example of the "inter-universal operations" discussed later, and it guarantees that the essence of a definition is not lost even when it spans multiple languages or representation systems.

Moreover, the hierarchical state grid is expected to play a critical role in these interlingual operations within the IMU. Since the state grid does not depend on the nature or domain of the system being described, one can consider how a conceptual state's coordinate in one system would map to coordinates in another. This allows the concept to be translated between languages immediately and with minimal friction. Furthermore, the hierarchical state grid itself is a meta-concept independent of any specific logic. Therefore, the very conceptual framework of "coordinatizing states in a hierarchical fashion" should, by its nature, be housed in the IMU. The IMU thus also plays the role of providing a place where such meta-concepts explicitly reside.

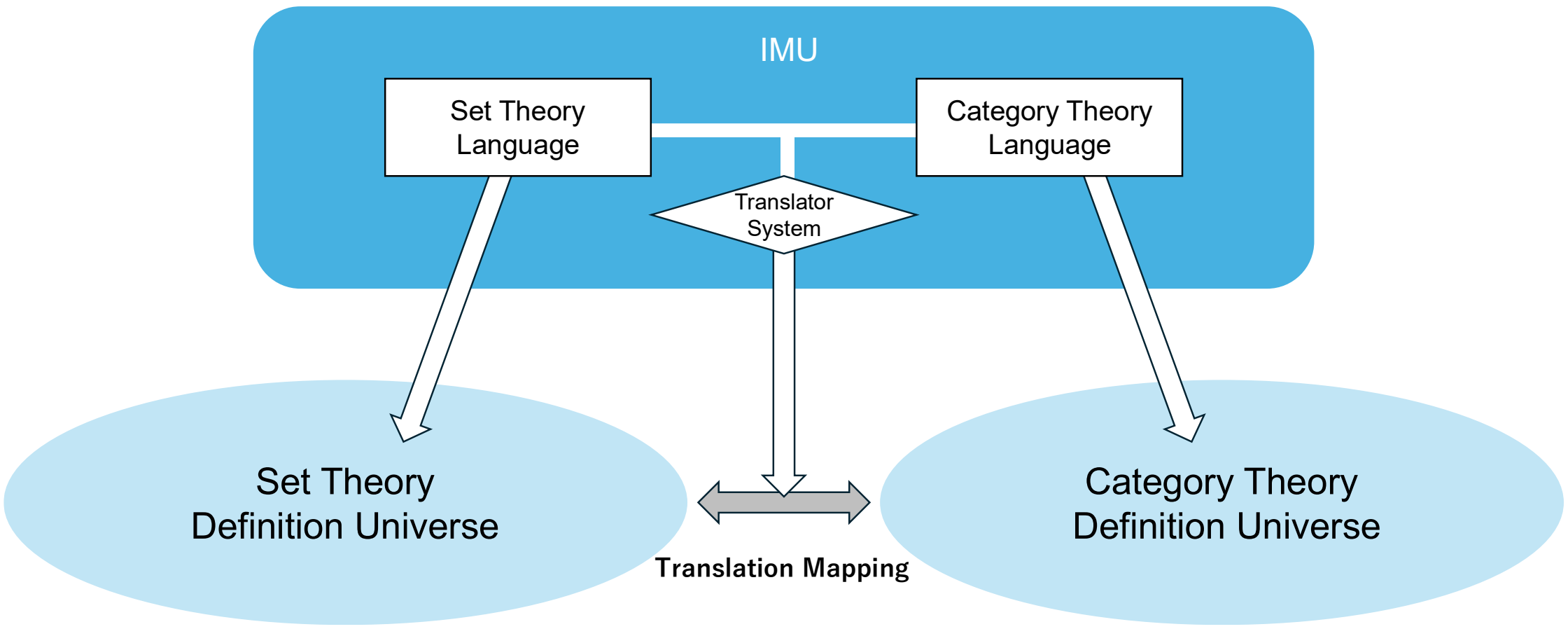

**Figure 2.** Image of translation mapping construction via IMU. Language definitions exist within the IMU, and framework-level meta-concepts for the definition universe are defined from the IMU. The content of translation mappings is also defined by retaining translation definitions within the IMU.

## 2. Defining Agents (Figure 3)

A multi-agent situation in which different agents—such as human A, human B, or artificial intelligence agents—edit and observe the definition universe from their own respective perspectives and methodologies is inevitable. Each agent possesses an independent observation point, editing method, and history, and it is likely that conflicts or collisions in state definitions frequently arise. To address such issues, the IMU provides an *integration mapping*. This serves to integrate edits and differences made by multiple agents safely and consistently, akin to three-way merges in Git [16] or automatic conflict resolution in CRDTs (Conflict-free Replicated Data Types) [17]. For example, by referencing the exact coordinates on the hierarchical state grids of each defining agent's universe, non-conflicting portions can be automatically integrated. At the same time, conflicting regions may remain as undefined states. This enables a clear and visual distinction between aligned and misaligned states. Moreover, since the defining agents themselves are explicitly represented as entities within the IMU, these multi-agent processes can be handled in an explicit and self-aware fashion. This ensures that even in environments involving multiple agents, *isomorphic state comparison* and *knowledge integration* are theoretically guaranteed. Such a framework is undoubtedly helpful for human agents, but it becomes especially indispensable for meta-level communication between artificial intelligences. Given the current absence of human-level empathy or ambiguity resolution capabilities in artificial intelligence, any meta-level communication between AIs demands formalization and explicit awareness of its methodology. Therefore, the introduction of a meta-logical layer such as the IMU is a requirement for mechanical knowledge systems.

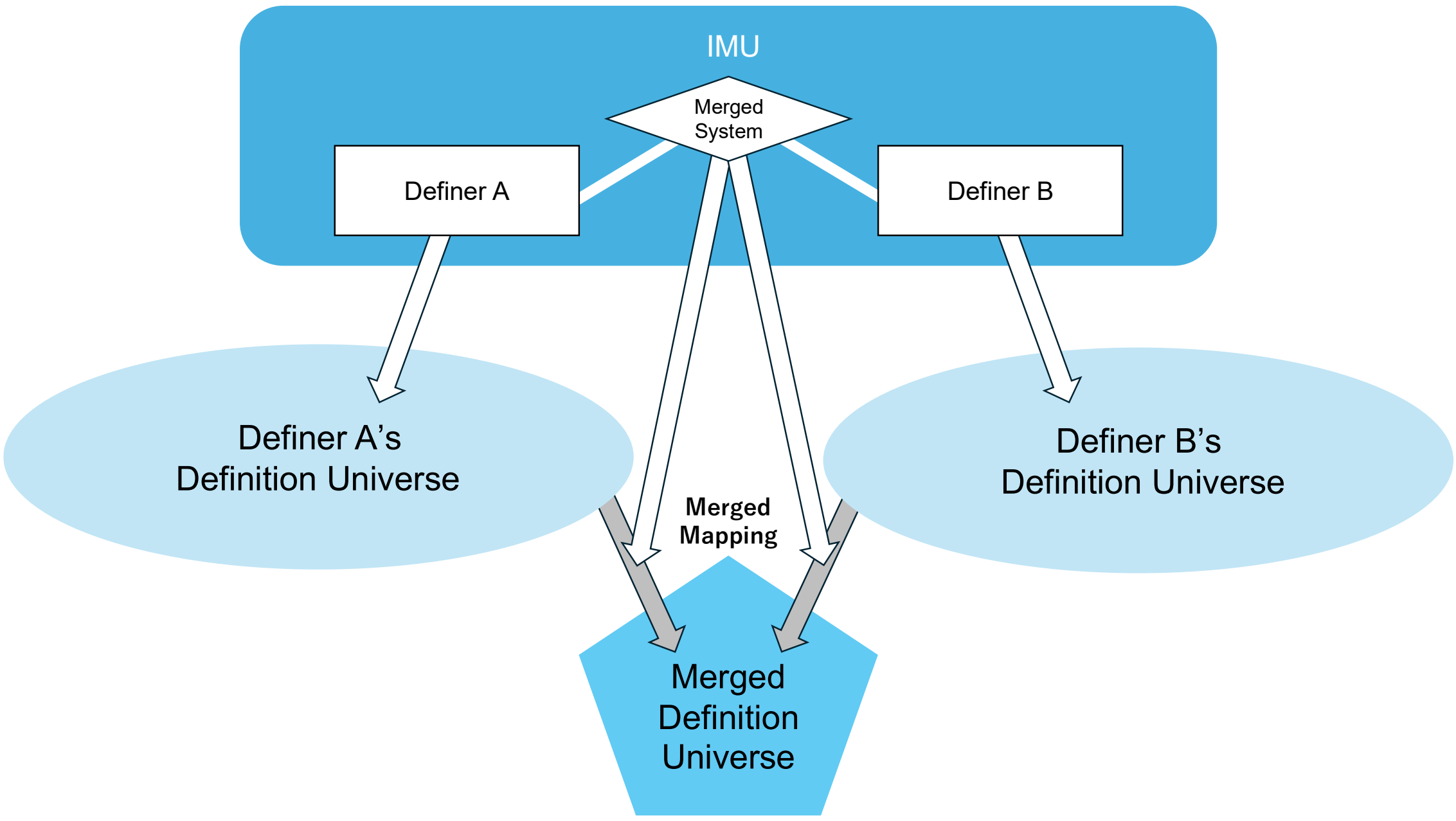

**Figure 3.** Image of constructing an integration mapping between agents via the IMU. Definitions of agent information within a given universe are retained in the IMU, from which an integration mapping is constructed through an agent-based integration system, forming a unified definition universe.

## 4. Real-Time Definition (Figure 4)

In the real world, the definition universe changes along with the progression of real time. In other words, any definition universe, when viewed from a meta perspective, is subject to the effects of real-time progression, and its contents are susceptible to change. To explicitly capture this phenomenon, the IMU provides real-time mapping. This refers to a mapping from a definition universe at time point $t$ to another at time point $t+1$, representing the transformation between them. Of course, as previously stated, it is fundamentally impossible to fully describe universe transformations at any point in the past, present, or future real-time progression. However, since the IMU forgoes any attempt at fully formalizing the true meta-universe, its mappings instead copy only a portion of the definition universe as time progresses. That is, if $t$ is taken as a past universe and $t+1$ as the present one, only observable or transferable subsets are mapped from the past to the present universe. Through this, unknown, unobserved, or lost information becomes *undefined*, and the relationships along the time axis can be clearly visualized.

Although it is impossible to access future information directly, the three-dimensional extension of the hierarchical state grid can serve as a pseudo-description mechanism for the future. For example, one can define a prediction as a state (including depth) on the 3D grid and later confirm whether that defined prediction was actually realized through real-time progression. In doing so, the grid comes to function as a framework for defining future states in a pseudo manner. This is precisely the intent behind placing definability at state depth zero as an axiomatic label on Table 2. In other words, placing "definability" at state depth zero serves to reserve within the theory a domain that is "not yet definable at present but may become definable in the future" as time progresses. For instance, some predicates may only be definable in a particular future moment during the course of real-time progression. The act of "verifying a prediction" is then nothing other than progressing in real time until the truth value of the prediction becomes definable—i.e., a predicate construction inherently tied to futurity (Of course, this also applies to predicates with existing truth values that are currently indefinable, or to predicates for which no truth value exists, as with historically lost events [truth exists but is undefinable] or logically unprovable propositions [truth does not exist]. These can all be meaningfully handled through the lens of definability.). In this way, predictive future states and unobserved regions can be explicitly managed and targeted within the grid via the IMU's real-time mapping. This, in turn, becomes the conceptual launchpad for clarifying the essential separation and distinction between "proof" (as a static means of validation) and "verification" (as a dynamic means), a topic further elaborated in the Inter-Universal Theory section.

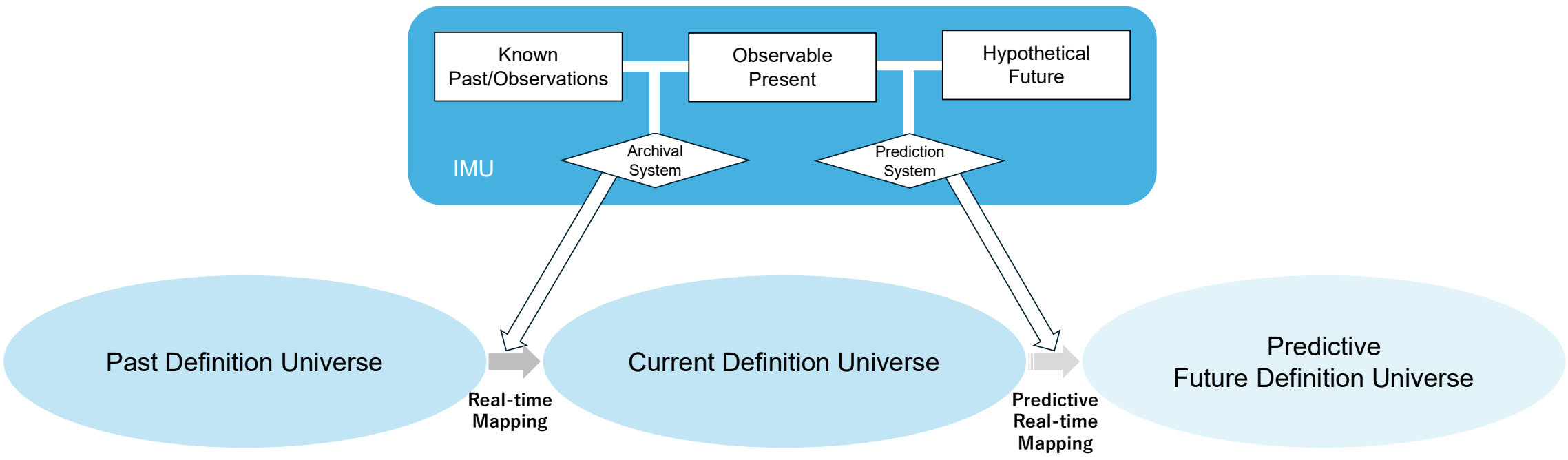

**Figure 4.** Image of real-time mapping construction via the IMU. Past, present, and future (predicted) information from a given universe is retained in the IMU, from which a real-time mapping is built based on record and prediction systems, thereby defining the universe's real-time progression.

Through this, the IMU acquires three primary functions—translation mapping, integration mapping, and real-time mapping—and becomes a central framework capable of fundamentally managing and controlling all forms of non-isomorphism across language, agency, and time. That is, regardless of what language a defined state is described in, which subject observes or modifies it, or at what point in time it is observed or verified, the structure of that definition or state is managed and unified in a consistent formalism. Accordingly, the introduction of the IMU provides a theoretically robust foundation for supporting the diverse and dynamic definitional operations of the real world—multilingual, multi-agent, and temporally evolving—without undermining the core principle of this study: “Definition = State.”

**Inter-Universal Theory in the Definition Universe**

The development of meta-universe theory introduced in this study naturally leads to the necessity of "inter-universal operations," which handle transformations across different universes—axiomatic systems, languages, agents, time, and more—in a unified manner. That is, if we aim to treat states and definitions in a genuinely universal way, it becomes inevitable to operate across these heterogeneous universes explicitly. Such inter-universal operations are likely to exist in various forms, extending beyond conventional purely mathematical operations. For instance, one can broadly distinguish between two categories: *macrocosm-inter-universal operations* and *microcosm-inter-universal operations*. "Macrocosm-inter-universal operations" refer to the transformations that span multiple language systems or axiomatic frameworks, preserving or transforming global and universal structures or concepts. These include operations such as "Change of Universe" in category-theoretic cosmology/universe theory [e.g., 10], which globally transfers entire theoretical frameworks, or the structural deformations seen in IUT [11], where the regular structure of a universe is altered. The former resembles equivalence transformations between universes, while the latter entails distortions in the mappings. These types of operations are widely used in contemporary mathematical logic and pure mathematics. On the other hand, "microcosm-inter-universal operations" refer to localized transformations that map only specific parts of a universe—such as particular concepts or structures—into another theoretical universe. This is often seen in practical, real-world scenarios outside mathematics, such as translating or converting technical terms or domain-specific concepts between different fields or language systems.

In traditional mathematics, inter-universal theory has mainly dealt with "structure-preserving transformations," often based on category theory and model-theoretic mathematical structures, or "structure-deforming transformations" as in IUT. These mathematical inter-universal theories are specialized in handling changes in size and properties of internal structures within axiomatic systems. However, by employing the meta-theoretical framework introduced in this study—centered around the IMU—it becomes possible to explicitly incorporate not only these mathematical inter-universal theories but also the previously mentioned non-mathematical inter-universal cases such as language translation, agent integration, and real-time transformation. These non-mathematical inter-universal theories generally exhibit microcosmic inter-universal characteristics. By explicitly distinguishing and managing such microcosmic inter-universality within the theory, this framework becomes applicable to scientific theories and logical systems of the real world beyond mathematics. In other words, this approach extends the concept of inter-universal theory itself. It serves as a vital foothold for developing a universal operational principle applicable even in the broader scientific domain.

Furthermore, the generalization of inter-universal theory leads to a clear separation between "proof" and "verification". Macrocosmic inter-universal logic is designed so that its logical structure is minimally affected by real-time progression. Therefore, in pure mathematics and ideal formal logic, legitimacy can be secured through proof. In contrast, microcosmic inter-universal logic, as discussed earlier, is inherently influenced by real-time progression—or at least integrates such influence into the theory. As such, it plays a central role in describing the verification process in non-mathematical fields such as science, engineering, and the real world (conceptually analogous to Karl Popper's falsifiability [18]). By introducing inter-universality in this way, we can clearly distinguish between the absoluteness and universality of proof in pure mathematics and the relativity and dynamism of verification in real-world contexts. This distinction provides a critical perspective for organizing and integrating the criteria for theoretical validity across modern science as a whole.

Finally, in inter-universal operations, the nature of the codomain (i.e., the theoretical destination) of each transformation becomes crucial. In all inter-universal transformations, the ultimate destination must theoretically be a universe where things are provable in the formal sense, provable in a form that can be shared/replicated, or verifiable in real-time progress. For instance, when a mathematical proof involves ambiguous natural language, it should not be left vague but converted into a machine-verifiable formal language (e.g., proof assistants such as Coq [12] or Lean [13]) to eliminate such ambiguity. Likewise, naive definitions (e.g., natural language-based or intuitively vague descriptions) or the axiomatization of such naive definitions as in Itoh (2025) should ideally be transformed into explicitly and concretely verifiable forms such as programming languages or simulation environments. Thus, this requirement for codomain ensures not only the theoretical soundness of the transformations but also provides the essential condition for knowledge and information to be meaningfully and safely shared across different universes.

In this way, the generalization of inter-universal theory evolves beyond being a mere mathematical technique. It becomes a theoretical and practical foundation across science as a whole, further reinforcing the framework of this study, which meta-theoretically integrates definitions and states.

# A Formal Framework for the Definition of 'State': Hierarchical Representation and Meta-Universe Interpretation

Kei Itoh*

## Abstract

本研究は、従来概念の定義に対して統一的な合意のないまま用いられてきた"状態"概念に対し、数学的かつ統一的な形式構造を導入することで、知能の公理的定義を含む多様な理論構築の基盤を補強することを目的とする。まず、"状態深度"と"写像階層"の 2 軸からなる"状態階層グリッド"を導入し、数理・物理・言語などあらゆる分野の状態を統一的な表記体系上で扱う手段を確立する。次に、我々（定義者）自身や使用言語自体を明示記述可能とする"中間メタ宇宙"を導入し、自己言及や不完全性を回避した形での自覚的なメタ的定義操作を実現する。さらに、このメタ論展開を基に、宇宙際理論を数学分野に留まらず、言語変換や主体統合など非数学的領域にまで拡張し、"大宇宙際／小宇宙際"の概念的区別を導入することで、広範な操作の記述を可能とする。主な成果をまとめると次の通り：

1. 分野間に非同型で曖昧だった"状態"概念を形式化し、定義・状態・構造を同一概念として統合。
2. 状態階層グリッドにより物理・論理・言語を問わず全状態を(状態深度, 写像階層)座標で表記。
3. 最下層に"定義可能性"を公理化し、未観測・未来事象を理論内に保持。
4. 述語・命題・プログラムをグリッド上で構造可視化し、依存関係の一望可能化。
5. 知能判定をブール関数化し、AI 実装に直結可能な評価写像として再構築。
6. グリッドを時間次元に拡張し、"仮想時間／実時間"を厳密に分離。
7. "中間メタ宇宙"で自己言及矛盾を遮断し、定義者・定義言語・定義のための定義を操作対象化。
8. 中間メタ宇宙に"翻訳写像"・"統合写像"・"実時間写像"を導入し、多言語性・多主体性・実時間変化を統合的に管理。
9. 宇宙際操作を"大宇宙際／小宇宙際"に二分し、非数学領域へ拡張。
10. 時間座標を用いて"証明（静的）"と"検証（動的）"の差異を明示。
11. すべての定義を検証可能宇宙へ輸送する終域の要請を設定し、形式化義務を明確化。

これらの構成により、本論文は知能定義・形式論理・科学理論全般に対して、"定義＝状態"という統合原理をもとに、時間・言語・主体・操作体系を貫きかつ科学全般へ適用可能なメタ数理論理に基づく枠組みを提示するものである。

* Ehime University Graduate School of Science and Technology. The author conducted this research independently while affiliated with Ehime University and utilized the university's library and electronic resources. The research was conducted without specific financial support from external funding sources. The author declares no conflict of interest.

**イントロダクション**

Itoh, 2025 [1]は、集合論もしくは圏論的宇宙を構築した上で、”知能”の公理的な定義を試みた。この研究は本来素朴的である概念に対して数学論理を導入可能にするプロトコルを実装し、実際に集合論論理による様々な知能の公理的な定義を行った。ここで、”要素”や”（時間）写像”といった集合論論理に基づく様々な概念が知能定義に対して導入されているが、その中でも”状態”という概念についてはその定義や操作方法について論文中で明示されていない。そして加えて、そもそも基礎的な集合論論理においてすらも”状態”という概念は明示的に何をどのように定義されるのか、未だに陽とした合意形成のない、曖昧な語彙として使われ続けてきた。例えば、制御理論の実数ベクトルとしての状態［e.g., 2]、C*-代数の正汎関数としての状態［e.g., 3]、モノイド圏での射としての状態［e.g., 4]、およびコーアルジブラでのキャリア集合要素としての状態［e.g., 5］など、実際に分野ごとに定義が相互に非同型である。このような状態に対する曖昧さは、概念に対して明示的でかつ数学的に厳密な操作を要求する公理的知能定義にとって致命的な欠陥である。したがって本論では、まず、これら分野間の”状態”概念の定義的断絶をその再解釈を通じて架橋し、知能の形式的定義の基盤理論を補強することを目的とする。そのために横軸に状態深度、縦軸に写像階層を取る”階層状態グリッド”を導入し、ここにおいてどのような数的、物理的もしくは他分野的なラベル・関係・写像など、ありとあらゆる”状態”と呼称されうるものを一括に同形式上で表記可能とすることを目指す。またこの構築では、”定義”という概念も”定義写像”というある一種の状態として見做せることにより、ある対象の”状態の決定”と”定義の決定”が実質的に同一視されることが示される。この展開により、”状態”と”定義”という論理上最重要な二つの概念が統合される。また、このように状態を階層として定義することにより、”時間”概念が、ある定義において時間的状態として導入される”仮想時間”と、我々自身にも常に作用し続けている時間である”実時間”という二つに分かれていて、その二つは明確に異なる概念対象であることが明らかになる。つまり、仮想時間は表記可能な数構造的非実在的時間であり即ちある形式による表記を用いて直接に定義可能であるが、実時間はメタ的に高度な概念であって直接に完全な定義は後述の理由から不可能であるものである、ということが状態階層の導入により鮮明になる。

以上をさらに包括的かつ円滑に操作可能なものとするため、本論ではこのような定義論に対して”定義者”や”定義言語”、”定義の為の定義”といったメタ概念対象を明示的そして自覚的に導入する。まず我々自身が存在する真にメタ的な宇宙と、定義を行う対象としての宇宙もしくは定義を内包する宇宙を階層として分け、前者を上階（真メタ宇宙）、後者を下階（定義宇宙）とする。また、本研究での”宇宙”は数学的宇宙のそれとして記述されていて、なおかつグロタンティーク宇宙［e.g., 6]のような特定の宇宙構成を示すものではなく、ある公理系によって記述されうる対象すべての集合、というより広い意味合いで使用される。ここで、我々自身という対象を完全に定義することは例えば不完全性定理［7]やラッセルのパラドックス［8]といった形式論的な不可能性から明らかに不可能である。もしくは

物理学的に考えてもパウリの排他原理 [9]を考慮すれば、我々自身を直接に完全表記するのは我々自身には不可能であると言える。そのため、本研究では真メタ宇宙と定義宇宙の間に表記上のバッファ的な役割として、”中間メタ宇宙”を導入する。中間メタ宇宙は真メタ宇宙に存在する対象の自己射的投射による鏡像的対象によって構築される宇宙であって、そして定義宇宙におけるメタ概念は中間メタ宇宙の対象の射によって定義付けられるとする。このようにメタ論を構築することによって、不完全性定理的完全定義不可能性を回避しつつ、定義におけるメタ概念を明示的に取り扱うことが可能となる。また、前述の実時間をどのように取り扱うかという問題に対してもこの考え方はその方法論を与えうる。

このメタ定義論は、Itoh, 2025 における知能の”素朴的定義”、”集合論的定義”、”圏論的定義”といった定義における使用言語的に異なる定義宇宙の、その操作方法の導入の動機を強く与える。すなわち、”実用性が担保される限り、定義はどのような言語でもよい”という主張は、本質的に宇宙際的操作方法の明示的導入を要請するものである。ここでいう”宇宙際的操作”とは、ホモトピー理論における”Change of universe [e.g., 10]”のような圏のサイズ制御やモデルの変換といった宇宙間変換や、宇宙際タイヒミュラー理論（IUT）[11]における”正則構造の変形”の論理構築といった宇宙間の性質の対応付けのような、宇宙間（≒公理系間）の数理論理的操作のことを指す。そして、Itoh, 2025 によって動機づけられて、本研究では、数学言語で記述されていない宇宙など考えうる全ての宇宙との宇宙際も包括した数理論理を導入したいと考える。すなわち、宇宙際という本来は純粋数学分野で扱われてきた概念を、科学一般の普遍的なものとして拡張することを目指す。またこのことは、特に Itoh, 2025 による知能定義といった言語間的宇宙際変換の”終域”は、その定義の真偽もしくは実用性を実時間進行において検証可能な宇宙でなければならないことを同時に示すことができる。この展開により例えば、ある証明完了が曖昧である数学論理はそれを機械証明可能な言語へ宇宙際変換し証明完了の検証をすべきであるという主張や、素朴的概念の公理的定義は最終的に機械的実装が可能な言語（プログラミング言語のような）へと宇宙際変換しその定義の実用性の検証をすべきであるという主張を、それぞれ自然に導くことができる。

従って本研究の目的は次の三つの論理構築である。

1. 状態階層グリッド——状態／定義を同一形式で扱う座標系
2. 中間メタ宇宙を主眼に置いた定義に対するメタ論——自己記述の限界を回避しながら定義操作をメタ外部化するバッファ層
3. 定義宇宙における宇宙際理論——言語・公理系を跨いで定義を検証可能な終域へ輸送する手続き

1 では、上述に加えて具体的な状態階層グリッドの定義・操作の方法を、関数連続性判定写像や Itoh, 2025 による知能定義をそれぞれ当てはめて定義することによって簡単に示す。その際に自然言語的集合論論理での構築と ZFC 系一階述語論理による定義構築を併せて記述する。加えて状態・定義の両概念が厳密化されることにより、Itoh, 2025 による”構造”の

定義の不足が浮き彫りになるため、これへ明示的な再定式化を行う。本論では 1-3 に通底して時間概念に対する状態論・メタ論的解釈を加えるが、その副次的帰結として、数学分野における”証明”という概念が、非数学分野において必ずしも論理の正しさを担保するわけではないということを示す。これは非数学分野においては全時間的に正しさが保証される操作というのが実質上存在しないのであって、理論の正しさは有限時間内の”検証”によってのみ担保されるということである。

**状態階層グリッド導入による状態定義論**

本研究が構築する"状態階層グリッド"とは、横軸を状態深度に、縦軸を写像階層に取る、様々な種類の"状態"を内包して定義可能な表記法である。状態深度とは、ある低次の状態から定義付けられる高次の状態があるとし、その高低を指す。どのような科学分野における状態も、ある単純な状態の複合によってある複雑な状態が定められるものであり、このことを一括に整理・表記するためにこのような概念を導入する。一方、写像階層とは、定義する写像的概念の定義域が写像であるか否かや、その階数もしくは次数（例えば高次圏論で導入されるような）を示す。0 階状態を非写像のラベル的状態とし、1 階写像はラベル的状態を定義域に持つ写像的状態、2 階写像は 1 階写像を定義域に持つ写像的状態…と、階層が上がるごとに写像的状態の次数が上がる。また表 1 に階層状態グリッドの簡単な概念的構成、表 2 に数体系状態を例にとった具体的なグリッド構築を示す。表 1 から、グリッドがデカルト座標表記によって座標化されることがわかる。またこの座標のそれぞれがある状態の構成を示している。

| 写像階 | | | | | | | |
|---|---|---|---|---|---|---|---|
| ⋮ | ⋮ | ⋮ | ⋮ | ⋮ | ⋮ | | |
| 3 階写像 | (0,3) | (1,3) | (2,3) | (3,3) | (4,3) | … | |
| 2 階写像 | (0,2) | (1,2) | (2,2) | (3,2) | (4,2) | … | |
| 1 階写像 | (0,1) | (1,1) | (2,1) | (3,1) | (4,1) | … | |
| 0 階状態 | (0,0) | (1,0) | (2,0) | (3,0) | (4,0) | … | |
| | 深度 0 | 深度 1 | 深度 2 | 深度 3 | 深度 4 | … | 状態深度 |

表 1.　階層状態グリッドの概念的構成。グリッドのそれぞれの座標が（横,縦）=(状態深度,写像階)を示している。

| 写像階 | | | | | | |
|---|---|---|---|---|---|---|
| 3 階写像 | (0,3) | (1,3)高階合成真理関数 | (2,3)濃度に関する高階性質 | (3,3)順序的高階性質 | (4,3)体の高階性質 | (5,3)連続体の高階性質 |
| | | | 全単射性判定、無限濃度判定 | 帰納法述語 | 代数的閉体判定、体演算の連続性判定 | 中間値定理、Banach 空間性判定 |
| 2 階写像 | (0,2) | (1,2)合成真理関数 | (2,2)濃度述語 | (3,2)順序述語 | (4,2)体の性質述語 | (5,2)連続性述語 |
| | | 論理和・積 | 濃度同一性判定、同型・可算性の述語 | 極大元、最小元 | 体の位相化可能性、距離化可能性 | 連続性判定、微分可能性判定 |
| 1 階写像 | (0,1)定義可能か判定する写像 | (1,1)真理値関数 | (2,1)濃度写像 | (3,1)順序構造写像 | (4,1)体写像／演算写像 | (5,1)連続性写像 |
| | | 論理ゲート | 可算、非可算判定 | 順序＋Peano 加算 | 四則演算 | 完備な関数 |
| 0 階状態 | (0,0)定義可能性 | (1,0)空/非空 | (2,0)基底集合 | (3,0)順序集合 | (4,0)体／環集合 | (5,0)完備体／実数体集合 |
| | 定義可能か否かの状態（メタブール） | ブール値 | 非順序基底 | ℕ、ℤ、全・半順序を持つ集合 | ℚ、加・乗法を持つ代数的数集合 | ℝ、完備性を持つ連続体 |
| | 深度 0 | 深度 1 | 深度 2 | 深度 3 | 深度 4 | 深度 5 |

状態深度

表 2.　階層状態グリッドによる数体系状態の表記。状態深度が深いほど複雑な概念の状態を示していて、写像階層が高いほどより述語的な状態を示す。写像階層はどのような写像的概念に対しても同型的に決められる可能性が高いが、状態深度は同じ状態概念系に対してであっても深度の取り方に任意性がある。

表 2 に示されているように、状態階層グリッドを導入することにより低次-高次状態と写像階数をすべて一括に表記可能であることが分かる。そして状態深度であれ写像階層であれ、その具体的内容は数学的・物理学的または他分野的のいずれかであることを制限しないために、この表記法はどのような分野の状態概念をも内包して定義可能であり、比較可能であると言えるだろう。そしてこれは表記法そのものであるために、状態深度・写像階ともにその具体的な内容は各人が定義したい状態によって変化するだろう。またここで重要なのは、ラベル的非写像状態だけに限らず写像概念をも状態だと捉え、そうすることでラベル概念

と写像概念を状態として統一していることである。こうすることで述語を状態と見做すことができるだけではなく、例えば関数空間系といった状態概念が写像性を内包する数理系をも内包可能となる。そして状態概念の一般的なイメージとしてのラベル的状態だけではなく、どのような高階述語をもある一つの状態と見做している構成であるがために、この表記法においては”定義の決定”と”状態の決定”が同一視され、定義概念と状態概念が統合される。

表 2 において深度 0 に関する状態を定義可能性やその可否に設定しているが、これはある定義を行う対象が定義可能であるか、という定義における根源的な状態を公理的に設定している。これはある定義したいものはそれを定義した途端にそれは定義可能なものとみることができるために、メタ視点を導入しない限りは全ての状態・定義に対して”定義可能”の状態を持つだろうことから根源的状態として導入すべき状態だと考える。つまり、深度 0 のブール値は、”その対象や状態が我々の記述系で”定義可能”であるか否か”という存在論的メタ判定であり、これは”空集合”ですらそう定義された時点で定義可能な状態としてカウントするものである。一方、深度 1 のブール値はすでに状態定義に載った対象に対する”属性や述語の真偽”であり、空か否か、成り立つか否かなどの内部的分岐を扱う。この記述だけでは深度 0 は設定する必要のない公理のように思われるがしかし、後に述べるように、実時間を考慮するとこの深度の区別は現実的な意味を持つ。我々は実時間進行において過去・現在・未来全ての時間の取り扱いにおいて、それが真に”定義不可能”な状態や述語に容易に遭遇するはずである。過去に存在した知ることのできない状態は確実に存在し、現在において知ることのできる状態は限られていて、未来の状態内容を今知ることはできない。つまり、現実の”実時間”を考慮すると、未観測・不可知・未来の事象が深度 0 で”定義不可能”として判定されて、その領域は理論的には無限に大きくなりうる。このように、定義論への実時間の導入を射程に入れると定義可能性は最も浅い状態として、ほとんどどのような状態概念へも公理的に導入すべき状態であると考える。

**状態階層グリッドを用いた具体的な状態定義**

**連続性判定写像の定義**

状態階層グリッドを用いた具体的な状態定義の例示として、まずは関数の連続性判定写像を定義する。以下の数式記述は ZFC 系一階述語を用いる。まず関数写像$f:\mathbb{R}\to\mathbb{R}$：

$$\mathrm{Func}(f)\equiv\forall x\forall y\forall z\big((x,y)\in f\wedge(x,z)\in f\to y=z\big),(\,x,y,z\in\mathbb{R}).$$

次に連続性述語$\varphi_{\mathrm{Cont}}(f)$：

$$\varphi_{\mathrm{Cont}}(f)\equiv\mathrm{Func}(f)\wedge$$
$$\forall a\in\mathbb{R}\forall\varepsilon\in\mathbb{R}^{+}$$
$$\exists\delta\in\mathbb{R}^{+}\forall x\in\mathbb{R}(|x-a|<\delta\to|f(x)-f(a)|<\varepsilon)^{(\mathrm{C})}.$$

ここで、

$$|x-a|<\delta\equiv(x-a<\delta)\vee(a-x<\delta)$$

などは順序$(<)$や体演算$(+,-,|\cdot|)$で展開可能である。そして$\varphi_{\mathrm{Cont}}(f)$をブール関数$\mathrm{Cont}(f)$を用いた真偽判定写像に変換する：

$$\mathrm{Cont}(f)\equiv\begin{cases}1 & \text{if}\,\varphi_{\mathrm{Cont}}(f)\text{が成り立つ}\\0 & \text{otherwise}\end{cases}.$$

ここで 0 = False, 1 = True とする。表 3 に示すように定義のために用いた以上の数式の内容は全て表 2 のグリッド上に配置できる。ここに示すように、ある定義の内容をすべて状態階層として書き下すとその状態が全て明らかになり、その構造が明示される。そのように状態階層グリッド上で定義の全要素を明示すれば、その定義が持つ本質的な構造・依存関係・操作深度が可視化される。

ここで、命題論理において命題やその真偽の記述は根源的には真理値関数の組み合わせによって行われ、そして計算機科学においてもどのような述語の記述も論理ゲートの組み合わせによって行われる。そのことから、ここまで論じてきた状態階層グリッドは、特に数学・情報科学的述語に関しては、どのような深さの状態もどのような高さの写像も、(1,1)真理値関数とその定義域である(1,0)空/非空の組み合わせによって展開されることが明らかに示される。

| グリッド座標 | ラベル | (C)式内で対応するパーツ |
|---|---|---|
| (5,0) | 実数体$\mathbb{R}$ | 変数$a,x,\varepsilon,\delta$の取りうる集合 |
| (4,1) | 体演算$+,-,|\cdot|$ | $x-a$, $f(x)-f(a)$の算術 |
| (3,2) | 順序述語"$<$" | 不等式 |
| (5,1) | 関数写像$f:\mathbb{R}\to\mathbb{R}$ | 関数写像を示す述語としての$f$ |
| (5,2) | $\varphi_{\mathrm{Cont}}(f)$ | 連続性述語 |
| (5,3) | $\mathrm{Cont}(f)$ | 連続性判定写像 |
| (1,0) | 真偽値{True, False} | 連続性判定の結果値 |

表 3.　連続性判定写像における各パーツの状態階層グリッド上（表２）の配置。

**Itoh, 2025 による公理的知能定義の再解釈**

次に、具体的な定義例としてもう一つ、Itoh, 2025 による公理的知能定義を状態階層グリッドを用いて再解釈する。そのためにまず、Itoh, 2025 における”要素”、”存在”、”構造”の定義を本論において明示化された状態概念を導入して再定義する。全宇宙集合$U$におけるその要素$v$は、ある状態$s$であるとする：

$$U \coloneqq \{v|\mathrm{Universe}\} \coloneqq \{s|\mathrm{Universe}\}$$

つまり、この宇宙は宇宙内に存在するある状態の全集合であると考える。宇宙中のある存在$E$はその部分集合であることは Itoh, 2025 と変わらない($E \subseteq U$)が、本研究による状態の導入により、それは同様に要素集合から状態集合へと再定義される。そして、ある存在$E$におけるある構造$C$は、直積集合を用いるのではなく、$E$におけるある特定の状態であると言える。これはつまり”構造”概念に対して独自の定義を行う必要が無くなったことを示す。状態階層グリッドにおいては、いかなる述語・関係式・写像的構成であっても、ある特定座標における状態として表現可能であり、それを”述語状態”と見做す。そのために、”定義と状態”が同一視されたのと同様に、”構造と状態”もまた同一視されるのである。実操作上は、ある構造はある座標のある状態として示され、その定義は具体的な定義に応じて行われるだろう。つまり本研究による再解釈として、構造とは、高階座標に現れる”述語状態”へ貼りつけられたラベルにすぎないことになる。この解釈は、例えば直積集合はそのような構造状態を”しばしば生成しやすい”道具立てであって、本質的な要件ではないということを示しているのである(他にも例えばべき集合や状態における絶対・相対の概念も構造状態の定義に用いられやすいと考えられる)。したがって本論では、定義・状態・構造は完全に同一概念の別称となり、あらゆる分野の”構造”概念を統一的に取り込むことが可能であることが示された。

続いて Itoh, 2025 における公理的知能定義を状態階層グリッドへと当てはめ、真偽判定述語として定義する。時間の進行を写像として概念化していることや”情報や物質を、外部から入力される構造、処理する構造、そして外部へ出力する構造のそれぞれを有する存在”が知能であるという素朴的定義、濃度演算と時間順序演算による述語はそのまま採用する。入力構造$C_i$、出力構造$C_o$、処理構造$C_p$はそれぞれ ZFC 系一階述語において以下のように記述される：

$$C_i \coloneqq |I_{i+1}| > |I_i| \wedge |O_{i+1}| > |O_i|$$

$$C_o \coloneqq |I_{i+1}| < |I_i| \wedge |O_{i+1}| > |O_i|$$

$$C_p \coloneqq \exists T, V \subseteq I(|T_{i+1}| < |T_i| \vee |V_{i+1}| > |V_i|)$$

そしてこれをブール関数$\mathrm{In}(I,O,i), \mathrm{Out}(I,O,i), \mathrm{Proc}(I,i)$へ変換する：

$$\mathrm{In}(I,O,i) \equiv \begin{cases} 1 & \text{if } C_i \\ 0 & \text{otherwise} \end{cases}$$

$$\mathrm{Out}(I,O,i) \equiv \begin{cases} 1 & \text{if } C_o \\ 0 & \text{otherwise} \end{cases}$$

$$\mathrm{Proc}(I,i) \equiv \begin{cases} 1 & \text{if } C_p \\ 0 & \text{otherwise} \end{cases}$$

ここで 0 = False, 1 = True とする。最後にこれを知能判定写像$\mathrm{Int}(I,O,i)$へと統合する：

$$\mathrm{Int}(I,O,i) \equiv \begin{cases} 1 & \text{if } \mathrm{In}(I,O,i) = \mathrm{True} \wedge \mathrm{Out}(I,O,i) = \mathrm{True} \wedge \mathrm{Proc}(I,i) = \mathrm{True} \\ 0 & \text{otherwise} \end{cases}$$

これで Itoh, 2025 による公理的知能定義は$\mathrm{Int}(I,O,i)$というブール関数として定義された。ここでの演算は濃度演算と順序演算に閉じており、表 4 のように表 2 の状態階層グリッドへ配置される。こうして純数学的概念だけではなく、知能のような非数学的概念もすべて高階述語状態に還元され、それは状態階層グリッドへ定義可能であることが示された。

| グリッド座標 | ラベル | 対応するパーツ |
|---|---|---|
| (2,0) | 基底集合 | $I, O$ |
| (2,1) | 濃度写像 | $\lvert I\rvert, \lvert O\rvert$ |
| (3,0) | 時系列順序集合 | $i, i+1$ |
| (3,1) | 順序写像（id, succ） | 時間写像$T_i: U_i \to U_{i+1}$ |
| (3,2) | 順序述語 | 不等式$\lvert I_{i+1}\rvert > \lvert I_i\rvert$など |
| (3,3) | $C_i, C_o, C_p$ | 知能構造述語 |
| (3,4) | $\mathrm{In}, \mathrm{Out}, \mathrm{Proc}$ | 知能構造判定写像 |
| (3,5) | $\mathrm{Int}$ | 知能判定写像 |
| (1.0) | 真偽値{True,False} | 知能判定の結果値 |

表 4.　知能判定写像における各パーツの状態階層グリッド上（表２）の配置。

**状態階層グリッドにおける時間概念の拡張**

Itoh, 2025 では時間は時間写像として定義され、それは実際に上述したようにある写像状態として記述可能であった。しかし時間概念を写像的に扱うことは本来的にはそれは代替手段であって、実際には時間は我々に直接作用し続けているメタ的概念であり、厳密あるいは実効上には我々が定義するもの自体よりも上位に存在すべき概念であるはずである。このことは状態階層グリッドの拡張として、座標の奥行きを時間の座標に割り当てることで表現可能であると考える（図１）。つまり、(横,縦)＝(状態深度,写像階)であった座標を、（横,縦,奥）＝(状態深度,写像階,時間)という３次元座標に拡張するのである。こうすると Itoh, 2025 における時間概念は写像ではなく状態階層グリッド上の座標へと還元される。そして時間写像を用いて知能を定義するのではなく、時間座標の異なる知能存在のその内容を比べることにより知能判定写像を構築するのである。その構築では表４のいくつかのパーツは必要がなくなり、加えてグリッド上に時間進行が明示されるのでより整理された状態定義が可能となるはずである。そして時間を独立した次元として取り扱うことは実時間と同型に時間概念を取り扱うということであり、後述する時間概念のメタ的操作の概念化、例えば”証明”と”検証”のその作用の違いを明示化することにも繋がる。

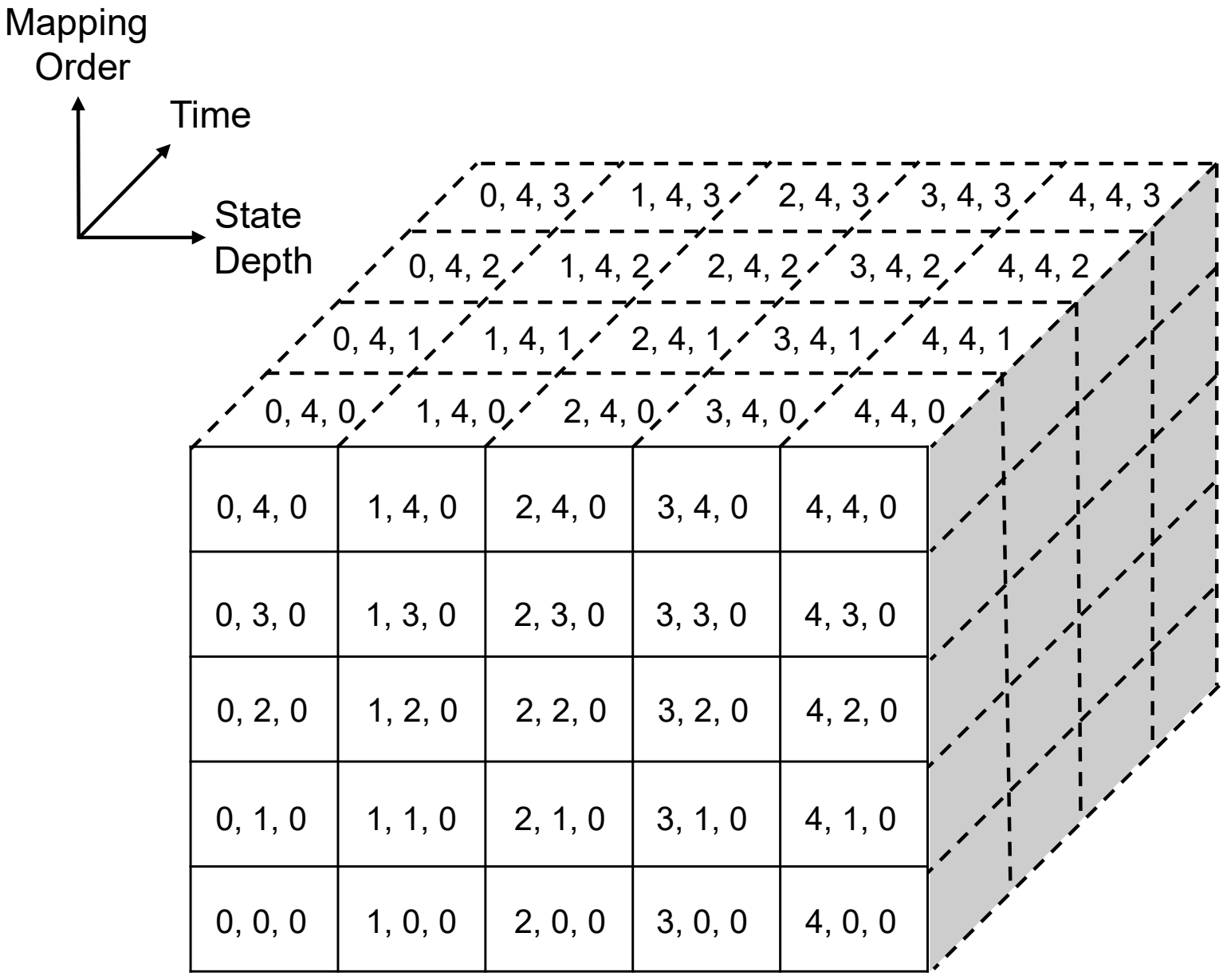

図 1. 時間拡張階層状態グリッドの概念的構成。グリッドのそれぞれの座標が（横,縦,奥）＝(状態深度,写像階,時間)を示している。

**中間メタ宇宙を主眼に置いた定義に対するメタ論**

本セクションでは、”中間メタ宇宙（IMU）”という概念の本質的意義と機能、ならびにその理論的・実用的な役割について詳述する。従来、どのような定義に関する理論体系でも、”我々自身”という観測者や我々自身が用いる言語や概念、そして我々自身に作用する実時間といった真にメタ的な対象（真メタ宇宙）は暗黙的に想定されている操作として扱われている。そして実際に記号体系や公理系、プログラム言語等の非メタの記述（定義宇宙）との二層構造も暗黙の前提としてきた。それは真メタ宇宙から定義宇宙へ直接的な作用や自己言及的な写像を行うことは、ゲーデル的不完全性や自己言及パラドックスを即座に引き起こし、理論体系そのものの健全性や一貫性を損なう根源的リスクを内包することに起因しているはずである。この自己言及の限界を回避するために不可欠となると考えられ、それ故に本研究で導入する概念が”中間メタ宇宙（IMU）”というバッファ層である。IMU は真メタ宇宙に含まれる対象の自己射的投射による鏡像的対象の集合として真メタ宇宙と定義宇宙との間に立つ。そして、定義宇宙におけるメタ的概念（定義者・定義言語・定義のための定義など）は、IMU からの射であると考えるのである。加えてこの概念を拡張し、IMU はすべての高次的な再帰操作、宇宙間写像、そして多様な情報変換・統合操作を安全に取り扱う場として機能することが示される。この層の本質は、自己言及矛盾や形式的な破綻を原理的に遮断しつつ、下位宇宙に対しては極めて柔軟かつ強力な”外部操作 API”を提供する点にある。実装レベルで言えば、各種メタ言語、機械証明言語［e.g., 12,13］、型理論でのユニバース多層構造［e.g., 14,15］などに対応しうる。つまり、この概念は特に情報科学的分野においては既にプログラミング言語の定義として暗黙的に取り入れられていて、それをより広範でかつ科学全般におけるメタ概念を内包するよう拡張する役割を果たすだろう。

IMU が対象化するメタ概念は、例えば以下の大きく三つのカテゴリに分けられる。

1.　**定義言語（図 2）**

一般にある定義において、集合論理論、圏論理論、さらにはさまざまなプログラム言語や自然言語のように、異なる構文・意味を持つ定義宇宙が同時並存することは普遍的に見られることである。そしてそれぞれの宇宙で同じ現象や構造を記述しようとしても、表記や定義形式は大きく異なりその統合方法の合意が存在しないために、それを実現しようと思うと現状では Itoh, 2025 の知能定義のように非数理論理的な職人的技法としてしか実装ができない。ここで IMU は、例えば”翻訳写像”という概念を提供する。つまり、IMU に各言語の”単語”や”文法”を対象として配置し、各言語間の対応関係を写像として対象化するのである。そしてこれを用いて定義宇宙間の翻訳関係を記述することにより、例えば集合論論理で定義された”状態”を、圏論や他の言語に数理論理として転写する機能を果たすだろう。もし変換不能や互換性の欠落が発生した場合、それは定義不能の定義として明示され、曖昧性や欠落をそのまま理論上で把握できる。これは後述する”宇宙際操作”の典型的な一例であると考えられ、そしてどのような定義言語・記述系にまたがっても定義の本質が失われないという保証をもたらす。また、状態階層グリッドはこの IMU における言語間操作においても重要な役割を果たすはずである。状態階層グリッドは前述のように記述する系の性質や分野を問わないために、ある体系における概念状態の座標が別の体系においてどの座標に配置されるかを考えることで、その概念は即座にそして容易に言語間翻訳され得ると考えられる。そしてそもそも、状態階層グリッドの考え方は特定の論理に立脚しないメタ的概念であるために、”状態を階層状に座標化する”という概念対象自体の置き場所は、IMU であるべきだと言えるだろう。IMU はそのようなメタ概念自体が明示的に存在する場所を提供する役割をも果たすのである。

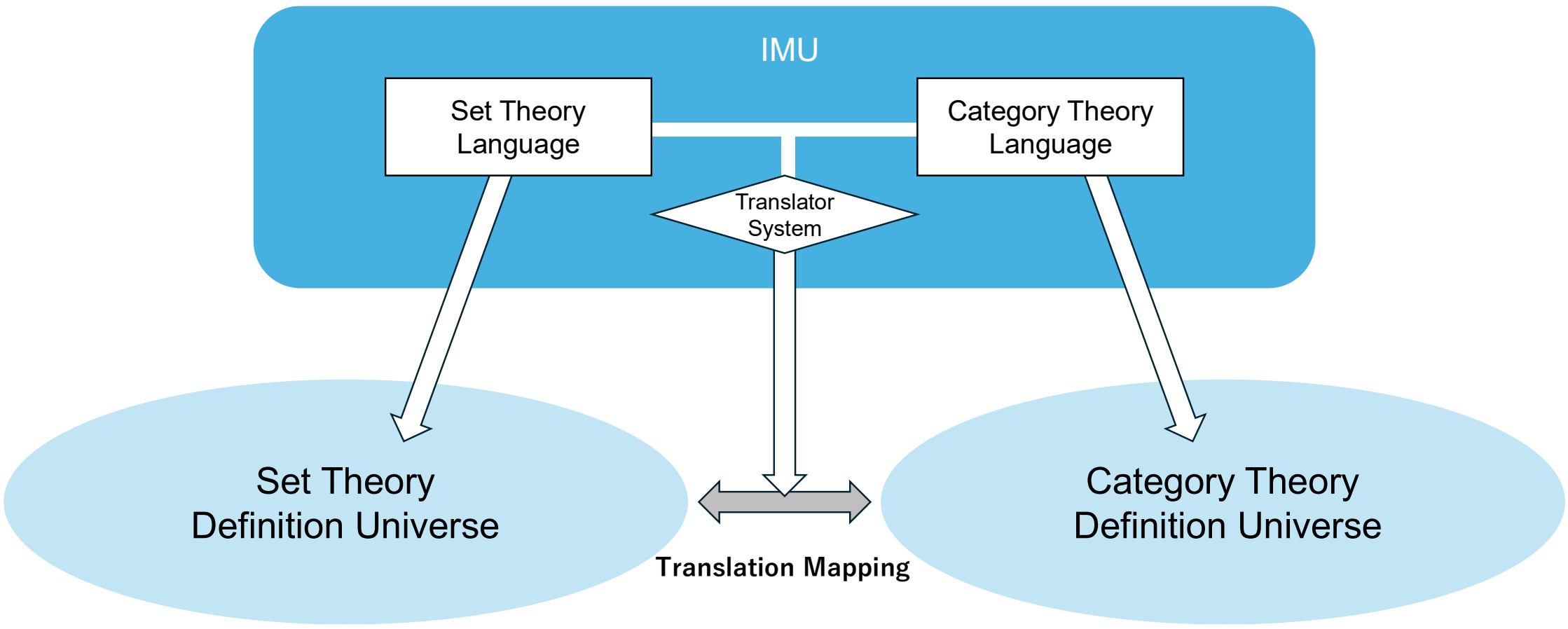

図 2．　IMU を介した言語間翻訳写像構築のイメージ。言語定義は IMU に存在し、定義宇宙のフレームワーク的メタ概念は IMU より定義される。翻訳写像の内容も IMU に翻訳定義を保持することにより定義される。

## 2.　定義主体（図 3）

　　人間 A・人間 B・人工知能エージェントなど、異なる主体がそれぞれの視点・方法で定義宇宙を編集・観測する多主体的な状況は不可避である。各主体は独立した観測点や編集方法・履歴を持ち、しばしば状態の競合や衝突も発生していると考えられる。このような問題に対処するために、IMU は”統合写像”を提供する。これは、Git の三方向 merge [16] や CRDT（Conflict-free Replicated Data Types）での自動 conflict 解消[17]のように、複数主体による定義の編集差分を、安全かつ整合的に統合する役割を担う。例えば、定義者宇宙間の状態階層グリッドの同座標状態を参照しながら、競合のない部分は自動的に統合され、競合箇所は未定義状態として残すなど、状態の一致・非一致を明確に可視化できるだろう。そして IMU に定義者自身が対象として存在することにより、このことを明示的・自覚的に取り扱うことが可能となるのである。これにより、多主体的な編集環境でも”同型的状態比較”や”知識統合”が理論的に保証される。こうした仕組みは人間主体にも有用だが、特に人工知能同士の定義参照やメタ的コミュニケーションには必須となる。人工知能に人間クラスの共感能力や曖昧性処理の能力が実装されていない現状、人工知能同士のメタ的なコミュニケーションはその方式の明示化と自覚化が必須であることは間違いない。ゆえに、IMU というメタ論理層の導入は、機械的な知識体系にとって不可欠な要件となるのである。

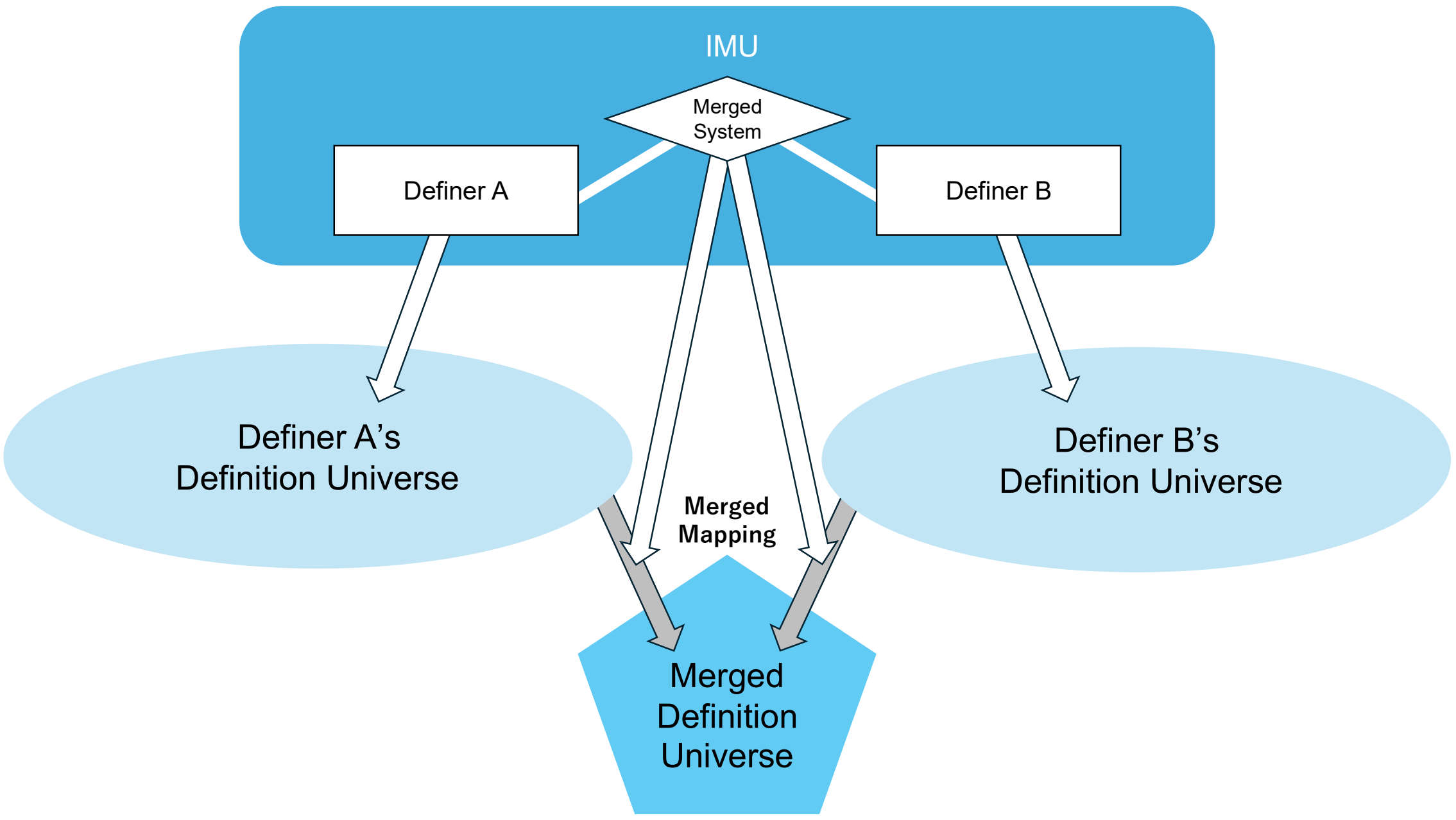

図 3.　IMU を介した主体間統合写像構築のイメージ。ある宇宙の定義者情報の定義を IMU に保持し、定義者間の統合システムから統合写像を構築し、統合宇宙を形成する。

## 4.　実時間定義（図 4）

実際の世界では、定義宇宙は現実の時間進行とともに変化する。つまり、どのような定義宇宙もメタ的に見れば実時間進行の作用を受けるのであって、その内容は容易に変化しうる。このことを明示化するために IMU は”実時間写像”を提供する。つまりこれは、ある時刻時点 t における宇宙と、その次の時刻時点 t+1 における定義宇宙を立てて、この間の変換としての写像である。もちろん、前述のように実時間進行における過去・現在・未来のいずれにおいても、宇宙変換を完全に記述することは不可能である。しかし IMU は真メタ宇宙の完全表記を放棄しているために、その写像は時間進行に伴い定義宇宙間のある一部のみをコピーする。つまり、t をある過去の宇宙、t+1 を現在の宇宙とすると、観測可能または引き継ぎ可能な部分集合のみを過去宇宙から現在宇宙へ写像する。こうして不可知・未観測・失われた情報は未定義状態となり、時間軸方向の関係を明確に可視化できる。未来の情報を直接取得することは不可能だが、特に 3 次元拡張の状態階層グリッドは実時間写像の疑似的な未来記述の役割を果たすことができるだろう。これは例えば、3 次元状態階層グリッドにおいて”予測”として奥行きを含めて状態定義し、そのある定義が実時間進行において実際にその予測通りに動作するかどうかを確かめることで、疑似的に未来状態の定義を担えることになる。そして、表 2 上に状態深度 0 の定義可能性を公理的に配置した狙いがここにある。つまり、表 2 で状態深度 0 に”定義可能性”を公理的ラベルとして置いたのは、実時間進行によって “現在はまだ定義できないが将来定義し得る” 領域を理論内に確保するためである。これは例えば、実時間進行過程においてある未来においてのみ定義可能である述語が明らかに構成できるのであって、そして”予測の検証”とはその予測の真偽が定義可能になるまで実時間を進行させるという未来性を孕んだ述語構成に他ならないのである（もちろん、過去・現在における真偽が存在するが定義不可能な述語や、真偽が存在しないために定義不可能な述語に対しても定義可能性の状態は有用に働く。たとえば、歴史的に証拠が失われた事象（真偽は存在するが定義不可能）や、論理的に証明不可能な命題（真偽が存在しない）においても、定義可能性という観点から管理することができる）。このように、予測的な未来状態や未観測領域も、IMU の”実時間写像”としてグリッド上で具体的に対象化・管理できる。そしてこれは、宇宙際理論セクションでさらに詳述する静的な正当性保証手段である”証明”と動的な正当性保証手段である”検証”の本質的な分離・区別を明示化するための主要な出発点でもある。

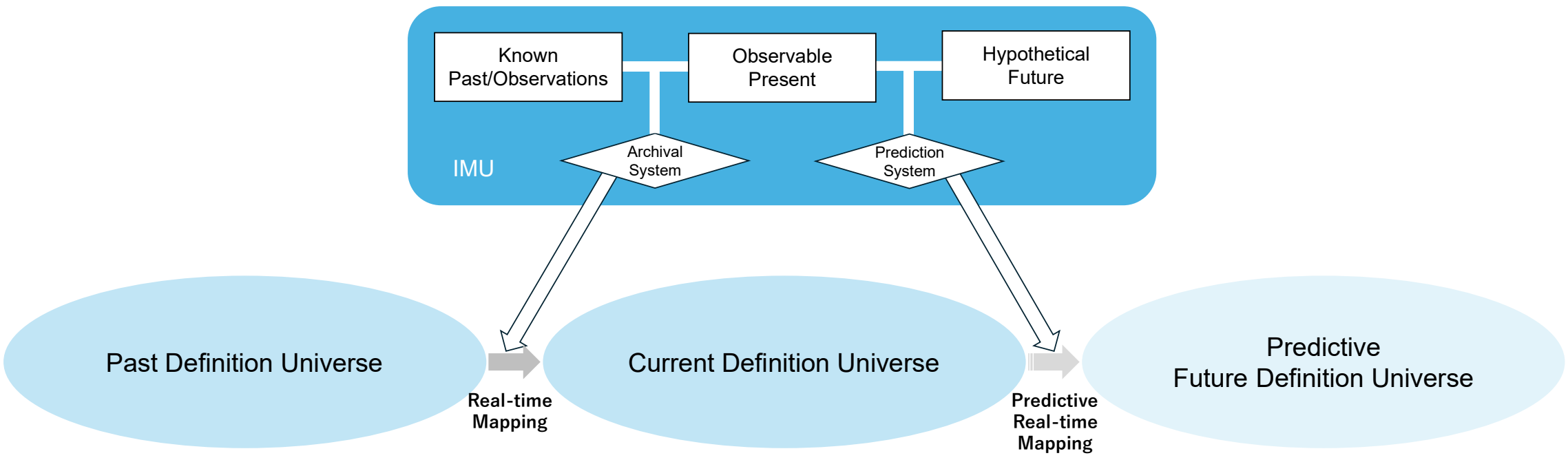

図４． IMU を介した実時間写像構築のイメージ。ある宇宙の過去・現在・未来（予測）情報の定義を IMU に保持し、時刻間の記録・予測システムから実時間写像を構築し、宇宙の実時間進行を定義する。

こうして IMU は、”翻訳写像”、”統合写像”、そして”実時間写像”という三つの主要機能を持ち得て、言語・主体・時間という全ての非同型性を原理的に管理・制御する中枢的枠組みとなる。すなわち、定義状態がいかなる言語によって記述されようと、どの主体によって観測・編集されようと、どの時点において観測・検証されようと、その定義や状態の構造は一貫した形式で管理・統合されることになる。したがって、IMU の導入は、”定義＝状態”という本研究の中核的な原理を崩すことなく、多言語的、多主体的、かつ時間的に動的な現実世界の多様な定義操作に対し、理論的で堅牢な基盤を提供する。

**定義宇宙における宇宙際理論**

本論で導入したメタ宇宙論の展開は、異なる宇宙間（公理系、言語、主体、時間など）の操作を統一的に扱う”宇宙際操作”の必要性を自然に導いている。つまり、状態や定義を真に普遍的に取り扱おうとすれば、必然的にそれらの異種宇宙間を横断する明示的操作が必要となるのである。こうした宇宙際操作には従来の純粋数学的操作から拡張されて多様な種類が存在するはずであり、例えば大きな区別に”大宇宙際（macrocosm-inter-universal）操作”と”小宇宙際（microcosm-inter-universal）操作”の二種類の大別が考えられる。”大宇宙際操作”とは、複数の言語体系や公理系を横断しかつ言語に全体的、普遍的な構造・概念の保存や変形の変換を指す。例えば圏論的宇宙論における”Change of Universe [e.g., 10]”のように理論そのものを全体的かつ普遍的に移動したり、IUT［11]のように宇宙における正則構造そのものを変形したりする際に用いられる操作であり（前者は宇宙の等価変換に近く、後者は写像の歪みを伴う変換である）、現代的な数理論理学や純粋数学において用いられている。一方で”小宇宙際操作”とは、宇宙のある限定された一部、つまりその宇宙に含まれる特定概念や構造だけを他の理論宇宙へ写像するような局所的な変換操作を指す。これは例えば特定の分野の専門用語や概念を他の分野や言語体系に翻訳・変換するといった、非数学として実務的かつ現実世界的な場面においてよく現れるだろう。

従来の数学分野では、宇宙際理論はもっぱら圏論・モデル理論的数学的構造を中心とした”構造保存変換”や、IUT のような”構造変形変換”を主眼として扱われてきた。こうした数学的宇宙際理論は、数学的公理系の内部構造のサイズや性質の変換に特化したものである。しかし、本研究が導入した IMU を代表する非数学にも通底しうるメタ理論の枠組みを用いることにより、このような数学的宇宙際理論だけでなく、前述したような言語翻訳・主体統合・実時間変換など非数学的な場面における宇宙際理論も明示的に取り込むことが可能になる。そのような非数学的宇宙際理論は一般に小宇宙際的な性質を持つと考えられるが、その小宇宙際性を理論において明示的に区別・操作する枠組みを整備することで、科学的理論や非数学的な実世界の論理体系まで広く適用可能となる。つまりこれは宇宙際理論そのものの概念を拡張し、科学一般においても応用可能な普遍的な操作原理へと発展させる重要な足掛かりとなりうる。

さらに、宇宙際理論の一般化は”証明”と”検証”の明確な分離を導く。すなわち、大宇宙際的論理はその論理構造が実時間進行の影響を受けづらいよう設計されており、したがって純粋数学や理想的な形式論理においては、”証明”による正当性の担保を強く保持することができる。一方、小宇宙際的論理は前述したように、常に実時間進行の影響を強く受けるか、少なくともその影響を理論に内包する性質を持つ。そのため、これは科学的・工学的・現実世界的な非数学分野における”検証”プロセス（概念としてはカール・ポパーの反証可能性［18]の考え方が至近）を記述する際に中心的役割を果たす。このように宇宙際性を導入することにより、純粋数学理論の”証明”における絶対性や普遍性と、現実世界での”検証”の相対性や動的性質を区別して取り扱うことが可能になる。この区別は、現代科学全般におけ

る理論的正当性の評価基準そのものを整理し統合するような重要な視点を提供するだろう。

最後に、こうした宇宙際操作においては、それらの変換の終域（定義として目指すべき最終地点）の性質が重要となる。すべての宇宙際変換操作においては、最終的に到達すべき終域は”証明可能”、”証明が共有可能”、もしくは”検証可能”な宇宙であることが理論的に要請されると言える。例えば、数学的証明が曖昧である自然言語を混交する数学理論による証明はそれを放置するのではなく、機械証明可能な形式言語（例えば Coq [12]や Lean [13]のような形式的証明アシスタント）へと変換することにより証明の曖昧性を排除することが望ましいと考えることができる。また、素朴的定義（自然言語的な定義や曖昧な直観的記述）あるいは Itoh, 2025 によるような素朴的定義の公理化は、ともに最終的にはプログラミング言語やシミュレーション環境のような明示的・具体的にその妥当性や正当性が検証可能な形式へと変換されるべきであると示唆できる。このように、終域の要請は理論の正当性を保証すると同時に、知識や情報が異なる宇宙間で有意義かつ安全に共有されるための必須の条件となるのである。

こうして、宇宙際理論の一般化は単なる数学的な技術的装置に留まらず、科学全般における理論的・実用的基盤へと発展し、”定義”と”状態”をメタ的に統合した本研究の枠組みをさらに強化するのである。